%% file: main.tex
\documentclass[a4paper,fleqn]{cas-dc}

\usepackage[authoryear,longnamesfirst]{natbib}

\def\tsc#1{\csdef{#1}{\textsc{\lowercase{#1}}\xspace}}
\tsc{WGM}
\tsc{QE}

\usepackage{amsmath}
\usepackage{amssymb}
\usepackage{amsthm}
\usepackage{multicol}
\usepackage{caption}
\usepackage{subcaption}
\usepackage{dsfont}

\begin{document}

\shorttitle{}    

\shortauthors{Y. Jeon et al.}  

\title [mode = title]{Mutually-Aware Feature Learning for Few-Shot Object Counting}  

\author[1]{Yerim Jeon}
\author[1]{Subeen Lee}
\author[1]{Jihwan Kim}
\author[1,2]{Jae-Pil Heo}
\cormark[1]
\affiliation[1]{organization={Department of Artificial Intelligence, Sungkyunkwan University},
            city={Suwon},
            postcode={16419}, 
            country={South Korea}}

\affiliation[2]{organization={Department of Computer Science and Engineering, Sungkyunkwan University},
            city={Suwon},
            postcode={16419}, 
            country={South Korea}}
\cortext[1]{Corresponding author at: Department of Artificial Intelligence, Sungkyunkwan University, Suwon, 16419, South Korea. E-mail addresses: jyr990330@gmail.com~(Y. Jeon), leesb7426@gmail.com~(S. Lee), damien911224@gmail.com~(J. Kim), jaepilheo@skku.edu~(J.-P. Heo)}

\begin{abstract}
Few-shot object counting has garnered significant attention for its practicality as it aims to count target objects in a query image based on given exemplars without additional training. 
However, the prevailing extract-and-match approach has a shortcoming: query and exemplar features lack interaction during feature extraction since they are extracted independently and later correlated based on similarity. This can lead to insufficient target awareness and confusion in identifying the actual target when multiple class objects coexist. To address this, we propose a novel framework, Mutually-Aware FEAture learning~(MAFEA), which encodes query and exemplar features with mutual awareness from the outset. By encouraging interaction throughout the pipeline, we obtain target-aware features robust to a multi-category scenario. Furthermore, we introduce background token to effectively associate the query's target region with exemplars and decouple its background region. Our extensive experiments demonstrate that our model achieves state-of-the-art performance on FSCD-LVIS and FSC-147 benchmarks with remarkably reduced target confusion.
\end{abstract}




\maketitle

\input{contents/1_introduction}
\input{contents/2_related_work}
\input{contents/3_method}
\input{contents/4_experiments}
\input{contents/5_conclusion}
\input{contents/6_appendix}






\clearpage
\bibliographystyle{cas-model2-names}

\bibliography{cas-refs}

\end{document}

%% file: contents/1_introduction.tex
\section{Introduction}
\label{sec:introduction}

Object counting has achieved remarkable advances along with deep learning networks.
However, most existing object counting methods are designed for a limited number of categories, such as human~\cite{macrowd} or car~\cite{car2}.
Those methods highly rely on a large amount of labeled data and cannot handle unseen categories beyond training data.
In this regard, few-shot object counting~\cite{GMN} has been proposed to count arbitrary class objects in a query image based on the given exemplar images.

Recent works~\cite{CounTR, LOCA} have explored both the few-shot and zero-shot settings.
The few-shot setting counts target objects based on exemplar images, while the zero-shot setting counts the most frequent class by identifying repetitive patterns without explicit guidance.
While both few-shot and zero-shot settings offer valuable approaches, few-shot setting plays a critical role in accurately defining the object unit in complex real-world applications. This is because recognizing repetitive patterns alone may not provide enough clarity regarding the unit of the object to count, whereas visual exemplars inherently contain information about the object unit.

A mainstream of few-shot object counting is the extract-and-match approach~\cite{GMN, CFOCNet, FamNet, BMNet+, RCAC, SAFECount, SPDCN, CounTR, LOCA, cstrans}.
Generally, this pipeline consists of three key components: 1) feature extractor, 2) relation learner, and 3) decoder.
Firstly, they compute query and exemplar features using the feature extractor, then construct the correlation volume through the relation learner. 
Afterward, they estimate the number of instances in the query image by transferring the correlation volume to the decoder.
\input{fig/figure1}

Although previous studies~\cite{SAFECount, LOCA} have achieved impressive performance, they exhibit a target confusion issue, failing to accurately identify only the target class when multiple classes of objects coexist in the query image, as shown in Figure~\ref{fig:motivation_fig}.
Existing methods have overlooked this problem, directly connected to the purpose of few-shot object counting, as benchmark datasets such as FSC-147~\cite{FamNet} primarily consist of single-class scenes.
The main reason for the target confusion is that the query features are computed without any explicit guidance of the target class.
Consequently, the query features tend to focus on objectness rather than target class-specific features, hindering the differentiation between target and non-target object features.

To address this, we propose a novel framework, Mutually-Aware FEAture Learning~(MAFEA), which enables the early consideration of mutual relations between query and exemplar features to produce the target class-specific features.
Specifically, MAFEA employs cross-attention to capture bi-directional co-relations between query and exemplar features, along with self-attention to reflect internal relationships.
With the cross-attention operating in a unified embedding space, the model can identify the difference between the target and non-target object features based on the exemplar features.
However, in the cross-attention, query background features, including non-target object features, are inherently expressed by exemplar features since there are no other features except exemplar features.
This might blur the distinction between the target and background features.
To prevent this, we introduce a background token, incorporated alongside exemplar features in self- and cross-attentions.
Our newly proposed Target-Background Discriminative~(TBD) loss trains this token to effectively represent background features, including non-target object features.
Consequently, MAFEA can capture the mutual relations beginning from the feature extractor and recognize the target objects clearly in the multi-category scenario, as shown in Figure~\ref{fig:motivation_fig}.

To sum up, our contributions are as follows:
\begin{itemize}
    \setlength\itemsep{5pt}
    \item To our knowledge, our approach is the first to tackle the target confusion issue, which involves accurately identifying the target class in a multi-class scenario.
    \item We propose a novel framework, Mutually-Aware FEAture Learning, which considers the mutual relationship between query and exemplar features from the outset.
    \item We introduce the background token and the Target-Background Discriminative loss to ensure a clear distinction between target and background representation. 
    \item Our method achieves state-of-the-art performance over baselines, and its effectiveness in a multi-class scenario is validated through comprehensive experiments.
\end{itemize} 

%% file: fig/figure1.tex
\begin{figure}[t]
    \centering
    \includegraphics[width=\columnwidth]{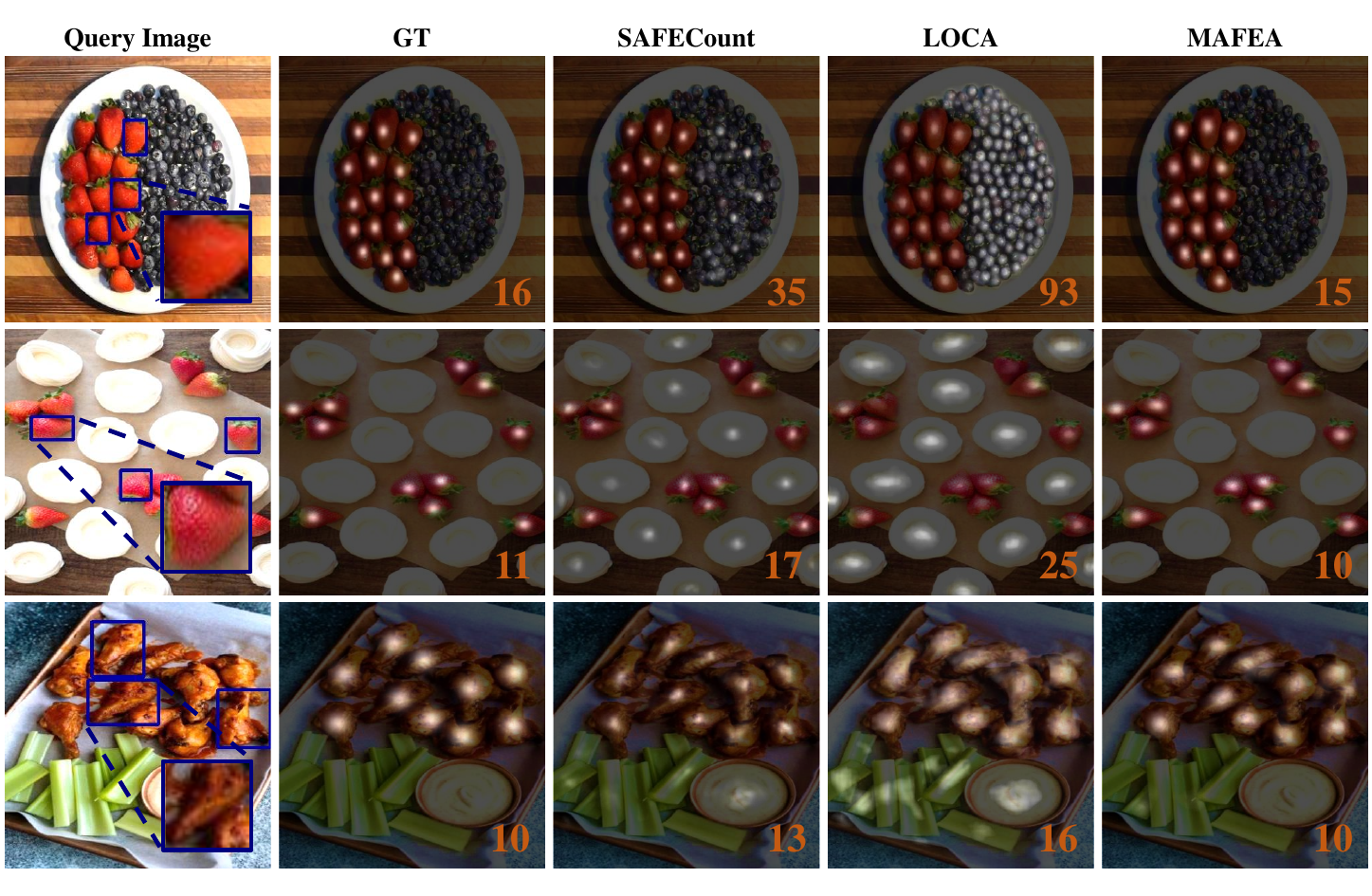}
    \caption{
    Target confusion problem in a multi-class scenario.
    Each box in the query image is a box annotation of an exemplar.  
    While SAFECount and LOCA count all objects in the query image regardless of the given exemplar images, MAFEA accurately distinguishes target objects based on the exemplars. 
    }
    \label{fig:motivation_fig}
\end{figure}

%% file: contents/2_related_work.tex
\section{Related Work}
\label{sec:related_work}

\subsection{Class-Specific Object Counting}
Class-specific object counting aims to count objects of a specific class, such as people~\cite{macrowd}, animals~\cite{animal}, and cars~\cite{car2}, in the images.
For this purpose, traditional methods~\cite{class-specific,class-specific2,class-specific3} solve the problem with detection-based approaches.
Since most datasets provide only point-level supervision, most detection-based methods generate the pseudo-bounding boxes from point-level ground truth and update them in the training phase.
However, they often struggle with scale variation and occlusion.
To alleviate this problem, regression-based approaches~\cite{class-specific-regression,class-specific-regression2, STNet, Interlayer} have emerged as popular alternatives in object counting, treating the task as dense regression to predict the object density map.
This approach adeptly tackles overlap problems and achieves good performance.
However, neither of these approaches can handle object classes not present in the training phase.

\subsection{Few-Shot and Zero-Shot Object Counting}

Few-shot object counting aims to count objects from arbitrary categories in a query image using only a few exemplar images.
Previous methods~\cite{GMN, CFOCNet, FamNet, RCAC, cstrans, BMNet+, fewshotseg, SAFECount, SPDCN, CounTR, LOCA} extract query and exemplar features independently, and match these features to infer a density map.
Recent methods such as CounTR~\cite{CounTR} and LOCA~\cite{LOCA} have evaluated zero-shot ability by replacing exemplar images with learnable tokens, extending beyond the few-shot setting. GCNet~\cite{GCNet}, on the other hand, proposed an exemplar-free approach and expanded it with weak supervision at the count level.

In parallel, zero-shot object counting has emerged, which enables counting based on text input instead of exemplar images. 
ZSCNet~\cite{ZSCNet} was the first to introduce a method that identifies objects of interest using class names.
More recently, SAM-Free~\cite{SAM-Free} and PseCo~\cite{PseCo} have attempted to leverage the powerful generalization abilities of foundation models like SAM~\cite{SAM} and CLIP~\cite{CLIP}.
However, since SAM is trained for segmentation and CLIP for classification, these approaches struggle in heavy occlusion, small objects, and dense scenes due to task discrepancies. 

It has been assumed that current few-shot and zero-shot counting methods can function in multi-class scenarios.
However, this has yet to be proved, as the benchmark dataset~\cite{FamNet} contains only one class per image. 
We experimentally confirmed that existing methods struggle with target confusion in multi-class scenes, whether in few- or zero-shot settings.
We conjecture that target confusion arises because existing methods encode query and exemplar features independently, without any explicit interaction during feature extraction. This lack of interaction causes insufficient target awareness, making it difficult for the model to distinguish between target and non-target objects in multi-class scenes.
To settle this problem, we designed the model to distinguish between target and non-target objects by producing target-aware features through early interaction in the feature extraction process. 
To the best of our knowledge, this is the first attempt to highlight and address the issue of target confusion.

%% file: contents/3_method.tex
\section{Method}
\label{sec:method}
\input{fig/figure2}

In this work, we introduce Mutually-Aware FEAture learning~(MAFEA) to compute query and exemplar features mutually-aware of each other throughout the feature
extractor, while the previous methods compute these features without any explicit feedback to each other, as illustrated in Figure~\ref{fig:main_figure_2}.
Specifically, MAFEA considers the co-relations between query and exemplar images in addition to the self-relations of each image.
Moreover, we introduce a learnable background token to prevent undesired interactions between the exemplar and background features in co-relations.
As a result, MAFEA can produce highly target-aware features that differentiate target object features from the background features, including the non-target object features.

\subsection{Overall Pipeline} 
The architecture of MAFEA consists of a ViT encoder, relation learner, and CNN decoder. Firstly, given a query image $I^Q\in\mathbb{R}^{3\times{H^Q}\times{W^Q}}$ and a set of $M$ exemplar images $I^E=\left\{I^{E}_i\in\mathbb{R}^{3\times{H^{E}_i}\times{W^{E}_i}} | i\in\left\{1, 2, \dots, M \right\}\right\}$ as input, they are split into $N^Q$ and $N^E_i$ image patches of resolution $S\times{S}$ respectively, where $N^Q=H^QW^Q/S^2$ for the query image and $N^E_i=H^E_iW^E_i/S^2$ for the exemplar images.
Then, image patches are converted into the query features $z^Q\in\mathbb{R}^{{N^Q}\times{C}}$ and exemplar features $z^E_i\in\mathbb{R}^{{N^E_i}\times{C}}$ by a projection function $\mathbb{R}^{3\times{S}\times{S}}\rightarrow\mathbb{R}^{C}$.
Also, position embedding is added to $z^Q$ and $z^E_i$ to retain positional information, and $z^E_i$ are concatenated to define $z^E\in\mathbb{R}^{{N^E}\times{C}}$ where $N^E$ denotes the sum of $N^E_i$.
After that, $z^Q$ and $z^E$ are refined by the ViT encoder, which incorporates mutual relation modeling.
Finally, the relation learner and CNN decoder sequentially receive the refined query and exemplar features and produce a density map $y\in\mathbb{R}^{{1}\times{H^Q}\times{W^Q}}$. The number of objects in the query image is computed as the sum of density map.

\subsection{Mutual Relation Modeling}
\label{subsec:MAFEA}
The core idea of MAFEA is to consider the mutual relationship between query and exemplar features from the outset.
MAFEA encodes features based on two types of relationships: self-relations within each image, and co-relations between different images.
The refined query $s^Q$ and exemplar features $s^E$, reflecting self-relations, are defined as follows:
\begin{eqnarray}
    \begin{aligned}
        \label{eq:self-relation}
        &s^Q = \text{MHA}(Q^Q,K^Q,V^Q), 
        \\
        &s^E = \text{MHA}(Q^E,K^E,V^E),
    \end{aligned}
\end{eqnarray}
where $Q$, $K$, and $V$ represent queries, keys, and values fed to the multi-head attention block~(MHA). 
We compute $Q^Q$, $K^Q$, and $V^Q$ by applying linear projections to the given query features $z^Q$ respectively, and produce $Q^E$, $K^E$, and $V^E$ using exemplar features $z^E$.
The self-relation modeling guides the query and exemplar features to capture self-similarity within each image.
Unlike the previous works only define self-relations throughout the feature extractor, we also refine query $c^{E\rightarrow{Q}}$ and exemplar features $c^{Q\rightarrow{E}
}$ using co-relations as follows:
\begin{eqnarray}
    \begin{aligned}
        \label{eq:co-relation}
        &c^{E\rightarrow{Q}} = \text{MHA}(Q^Q,K^E,V^E),
        \\
        &c^{Q\rightarrow{E}} = \text{MHA}(Q^E,K^Q,V^Q).
    \end{aligned}
\end{eqnarray}
The correlation modeling enables bi-directional interaction between query and exemplar features.
Firstly, the exemplars influence the query by identifying the difference between the target and non-target object features. 
Secondly, the query contributes to refining the exemplars, enabling them to aggregate diverse target object features. 
As a result, the encoder refines the query features to focus more on target-specific traits rather than general object characteristics.
With self-relations and co-relations, the output sequences of the $l$-th encoder layer are derived as follows:
\begin{eqnarray}
    \label{eq:encoder}
    \begin{aligned}
    &z^Q_{l+1} = z^Q_{l} + s^Q_{l} + c^{E\rightarrow{Q}}_{l},\quad l=1,2,\dots,L-1 ,
    \\
    &z^E_{l+1} = z^E_{l} + s^E_{l} + c^{Q\rightarrow{E}}_{l},\quad l=1,2,\dots,L-1 ,
    \end{aligned}
\end{eqnarray}
where $z^Q_{l}$ and $z^E_{l}$ are the input sequences of the $l$-th encoder layer. 
This modeling enables the encoder to adapt features based on their inherent self-relations and their interrelated correlations.

\subsection{Background Token}
When computing $c^{E\rightarrow{Q}}$ in Eq.~\ref{eq:co-relation}, MAFEA utilizes only exemplar features to produce keys and values.
Although the attention mechanism intrinsically mitigates improper co-relations, the exemplar features might represent the background features, including non-target object features.
In this case, it obscures the difference between the target object and background features; thus, it confuses the precise identification of the target in the query features.
In this regard, we introduce the background token, designed to learn the general features of the background region.
The background token $z^B\in\mathbb{R}^{1\times C}$ is concatenated with the exemplar features and then fed into the self-relation and co-relation modeling as follows:
\begin{eqnarray}
    \begin{aligned}
        \label{eq:bg-co-relation}
        &s^{[E;B]} = \text{MHA}([Q^E; Q^B], [K^E; K^B], [V^E; V^B]),
        \\
        &c^{[E;B]\rightarrow{Q}} = \text{MHA}(Q^Q, [K^E; K^B], [V^E; V^B]),
        \\
        &c^{Q\rightarrow{[E;B]}} = \text{MHA}([Q^E; Q^B], K^Q, V^Q),
    \end{aligned}
\end{eqnarray}
where $Q^B$, $K^B$, and $V^B$ are obtained by linear projections on the background token, respectively.
$s^{[E;B]}$, $c^{[E;B]\rightarrow{Q}}$, and $c^{Q\rightarrow{[E;B]}}$ substitute $s^{E}$, $c^{E\rightarrow{Q}}$, and $c^{Q\rightarrow{E}}$ defined in Eq.~\ref{eq:self-relation} and Eq.~\ref{eq:co-relation}, individually.
By incorporating the background token into those relations, 
we can prevent the background features from being expressed by the exemplar features in the computation of $c^{[E;B]\rightarrow{Q}}$.

\subsection{Target-Background Discriminative Loss}
Although the background token is designed to handle the background features of the query, it is not guaranteed without an explicit objective.
In this regard, we define a target-background discriminative~(TBD) loss, which encourages the background token to align with background features.
We first compute alignment score $AS_i$, which represents the degree of alignment between $i$-th query feature and the background token, as follows: 
\begin{equation}
    \begin{aligned}
        \label{eq:AS}
        AS_i=
        \frac{
        {\sum_{j=1}^{N^B}\left(\text{exp}\left(Q^{Q}_i \cdot K^{{B}}_j\right)\right)}
        }
        {\sum_{j=1}^{N^E+N^B}\left(\text{exp}\left(Q^{Q}_i \cdot [K^E; K^B]_j\right)\right)}
        ,
    \end{aligned}
\end{equation}
where $Q^{Q}_i$ is $i$-th $Q^{Q}$, and $K^{{B}}_j$ and $[K^E; K^B]_j$ denote $j$-th $K^{{B}}$ and $[K^E; K^B]$, respectively. 
Then, to align the background token only with background features, we divide the query features into positive set $P$, comprising features that spatially contain one or more ground-truth~(GT) points, and negative set $N$ which consists of features not including any GT points.
As a result, we define TBD for the $i$-th query feature, as follows:
\begin{equation}
    \begin{aligned}
        \label{eq:TBD}
        \mathcal{L}^{\text{TBD}}_i
        = -\mathds{1}_{z^Q_i\in P}\text{log}\left(1-AS_i\right) 
        - \mathds{1}_{z^Q_i\in N}\text{log}\left(AS_i\right),
    \end{aligned}
\end{equation}
where $z^Q_i\in P$ and $z^Q_i\in N$ mean $i$-th query features belong to the positive set and or not, respectively.
Also, $\mathcal{L}^{\text{TBD}}$ is the average value of $\mathcal{L}^{\text{TBD}}_i$ over all query features.

\subsection{Training Loss}
Once obtaining the target-aware features, $z^Q$ and $z^E$, we produce the correlation volume using the relation learner and convert it to the density map with the decoder. Following \cite{LOCA}, the relation learner performs iterative adaptation to produce intermediate correlation volumes, subsequently processed by auxiliary decoder blocks as follows:
\begin{equation}
    \begin{aligned}
        \label{eq:decoder}
        &C = \text{Relation-Learner}(z^Q, z^E), \\
        &y_{k} = \text{Decoder}_{k}(c_{k}), \quad k=1,2,\dots,K ,
    \end{aligned}
\end{equation}
where $C=\{c_k\}^{K}_{k=1}$ is the set of correlation volumes and $y_k$ is the output of the $k$-th decoder.

We adopt the object-normalized $l_2$ loss~($\mathcal{L}^{\text{count}}$), which is the mean squared error between the predicted and ground truth density map normalized by the number of objects. 
The object-normalized $l_2$ loss is formulated as follows:
\begin{equation}
    \begin{aligned}
        \label{eq:L_count}
        \mathcal{L}^{\text{count}} = \frac{1}{M}\left\vert\vert y - \hat{y}\vert\right\vert_{2}^{2},
    \end{aligned}
\end{equation}
where $y$ and $\hat{y}$ are the predicted density and ground-truth density maps, respectively. $M$ is the number of objects in mini-batch.
Also, we utilize the auxiliary loss~($\mathcal{L}^{\text{aux}}$) for the intermediate density maps as follows:
\begin{equation}
    \begin{aligned}
        \label{eq:L_count}
        \mathcal{L}^{\text{aux}} = \frac{1}{M}\sum_{k=1}^{K-1}\left\vert\vert y_{k} - \hat{y}\vert\right\vert_{2}^{2},
    \end{aligned}
\end{equation}
where $y_{k}$ is the intermediate density map of the $k$-th decoder and $K-1$ is the number of intermediate density maps.
The full objectives are defined as follows:
\begin{equation}
    \begin{aligned}
        \label{eq:full_loss}
        \mathcal{L} = \mathcal{L}^{\text{count}} +
       \lambda_{1}\mathcal{L}^{\text{aux}} +
        \lambda_{2}\mathcal{L}^{\text{TBD}},
    \end{aligned}
\end{equation}
where $\lambda_{1}$ and $\lambda_{2}$ are the weights of the auxiliary loss and TBD loss, respectively. 

%% file: fig/figure2.tex
\begin{figure*}[t!]
    \centering
    \includegraphics[width=\textwidth]{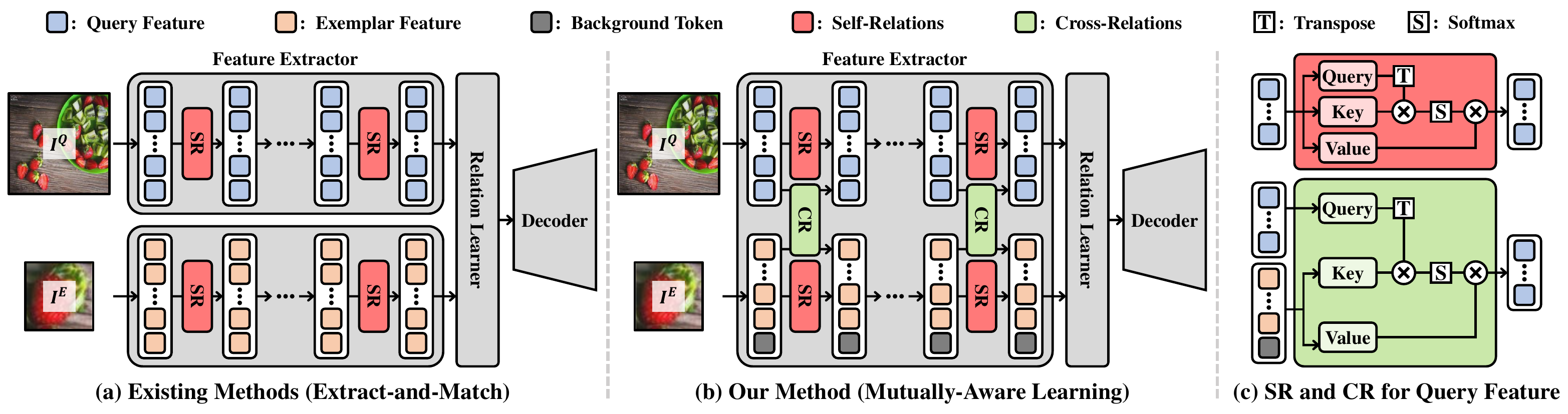}
    \caption{
    Comparison between Extract-and-Match methods and our proposed MAFEA.
    (a) Existing methods extract query and exemplar features without any explicit feedback to each other. 
    (b) On the other hand, MAFEA produces the query and exemplar features based on their mutual relation from an early stage of the feature extractor. By integrating self-relations and bi-directional co-relations, MAFEA produces highly target-aware features.
    Moreover, the learnable background token is fed into the self- and co-relations with the exemplar features to represent the background regions of the query image.
    (c) Self- and Cross-Relations are implemented by self- and cross-attention mechanisms.
    }
    \label{fig:main_figure_2}
\end{figure*}

%% file: contents/4_experiments.tex
\section{Experiments}
\label{sec:experiments}

\subsection{Implementation Details}
\subsubsection{Architecture.}
Our framework comprises a ViT encoder, relation learner, and CNN decoder. 
The patch size is set to $16\times16$, and both the kernel and stride of the projection head are set to $16\times16$ corresponding to the patch size.
The ViT encoder comprises 12 transformer encoder blocks where the hidden dimension of each block is 768, and the multi-head attention of each block consists of 12 heads.
The relation learner, inspired by LOCA~\cite{LOCA}, incorporates Object Prototype Extraction~(OPE) modules and Prototype Matching~(PM). OPE integrates object shapes and appearance properties using a three-layered iterative adaption module. Each layer includes two multi-head attention blocks and a feed-forward network.
Instead of ROI pooled features, we utilize exemplar features as appearance properties. PM involves the depth-wise correlation and max operation. Further details are provided in the appendix.
The CNN decoder comprises 4 convolutions and bilinear upsampling layers to regress a 2D density map.
The ViT encoder is initialized with a self-supervised pre-trained weight from MAE\cite{MAE}.
On the other hand, the parameters of the relation learner and CNN decoder are randomly initialized.

\subsubsection{Training details.}
We apply the same pre-processing and augmentation strategies as LOCA.
For pre-processing, the query image is resized to $512\times512$ and exemplar images are scaled to $48\times48$ based on provided box annotations.
Tiling augmentation is then applied, where the image is split into multiple tiles, and random horizontal flipping and color jitter are applied to each tile.
In the zero-shot setting, where no exemplar is provided, we follow the same approach as LOCA by replacing exemplar images with a learnable token for training and evaluation.
The weights $\lambda_{1}$ and $\lambda_{2}$ for auxiliary and TBD loss in Eq.~\ref{eq:full_loss} are set to $0.3$ and $0.05$, respectively. 
AdamW optimizer is employed with a batch size of 8, an initial learning rate of $1e^{-4}$, which is halved every 40 epochs, for a total of 100 epochs.
All experiments are conducted on a single RTX3090 GPU, with a training time of approximately 10 hours and an inference time of 0.02 seconds per sample. 

\subsection{Datasets and Metrics.}
\subsubsection{Datasets.}
We experiment on three benchmark datasets:  FSCD-LVIS~\cite{Counting-DETR}, FSC-147~\cite{FamNet}, and CARPK~\cite{CARPK}.
FSCD-LVIS is designed for few-shot object counting and detection in complex scenes with multiple class objects. 
It contains $6195$ images across $372$ classes, split into $3953$ train and $2242$ test images.
FSC-147, a few-shot object counting dataset, consists of simpler scenes where most images contain only a target class.
It includes $6135$ images of $147$ classes, divided into $3659$ train, $1286$ validation, and $1190$ test images. 
Both datasets include three randomly selected exemplars per image to depict the target objects. 
Note that, there is no shared object class between the sets.
Furthermore, we validate our models' generalization capability on the test set of CARPK, a dataset tailored for counting cars, comprising $459$ drone-captured images.

\begingroup
\begin{table}[t]
    \caption{Image indices of FSC-147-Multi. The subset consists of 31 and 12 images in the validation and test set, respectively.}
    \label{tab: multi-class subset indices}
    \centering
    \begin{tabular}{c|c}
        \hline\hline
        Eval Set & Indices \\
        \hline \hline
        \multirow{5}{*}{Val} & $216$, $236$, $243$, $244$, $252$, $752$, $913$, \\ 
                            & $1930$, $1999$, $2303$, $2305$, $2306$, $2826$,  \\
                            & $2830$, $2837$, $2868$, $2872$, $2875$, $2890$, \\
                            & $3520$, $3592$, $3785$, $3979$, $3980$, $4102$, \\
                            & $4851$, $5103$, $5105$, $5111$, $5669$, $6872$ \\
        \hline
        \multirow{2}{*}{Test} & $336$, $343$, $344$, $681$, $2143$, $3114$, \\
                            & $4495$, $4885$, $4920$, $4921$, $5379$, $6732$ \\
        \hline\hline
    \end{tabular}
\end{table}
\endgroup

\begingroup
\renewcommand{\arraystretch}{1} 
\begin{table*}[t]
	\caption{Evaluation on multi-class datasets in zero- and few-shot settings. The results on FSCD-LVIS are reproduced using the official implementation, except for Counting-DETR, which uses its official performance. The results on the multi-class subset of FSC-147~(FSC-147-Multi) are evaluated using the official pre-trained weights. The subset consists of 31~(over 1286) in the validation set and 12~(over 1190) images in the test set, totaling 43 images. `-' means that the score is not available and `$\times$' means that no exemplar is provided.}
	\label{tab: Multi-scenario table}
	\centering
	\begin{tabular}{l|c|cc|cc}
		\hline\hline
            \multirow{2}{*}{Method} &
            \multirow{2}{*}{Exemplar} &
            \multicolumn{2}{c|}{FSCD-LVIS} &
            \multicolumn{2}{c}{FSC-147-Multi} \\
            \cline{3-6} 
            & & MAE~$\downarrow$ & RMSE~$\downarrow$ & MAE~$\downarrow$ & RMSE~$\downarrow$ 
            \\ 
            \hline \hline
            SAM-Free~\cite{SAM-Free} & class name & - & - & $17.79$ & $26.74$  \\
            PseCo~\cite{PseCo} & class name & - & - & $14.74$ & $23.23$ \\
            \hline
            CounTR~\cite{CounTR} & $\times$ & - & - & $14.17$ & $28.80$ \\
            LOCA~\cite{LOCA} & $\times$ & - & - & $12.91$ & $24.39$ \\
            MAFEA~(Ours) & $\times$ & $14.72$ & $27.42$ & $12.12$ & $26.51$ \\
            \hline
            Counting-DETR~\cite{Counting-DETR} & box & $23.50$ & $35.89$ & - & - \\
            BMNet+~\cite{BMNet+} & box & $17.04$ & $27.71$ & $11.22$ & $16.99$ \\
            SPDCN~\cite{SPDCN} & box & $14.36$ &$26.31$ & $9.27$ & $13.54$ \\
            CounTR~\cite{CounTR} & box & $14.14$ &$26.01$  & $16.12$ & $30.70$ \\
            SAFECount~\cite{SAFECount} & box & $14.01$ & $24.04$  & $8.63$ & $12.18$ \\
            LOCA~\cite{LOCA} & box & $14.38$ & $24.28$ & $9.97$ & $16.83$ \\
            SAM-Free~\cite{SAM-Free} & box & - & - & $13.12$ & $18.20$ \\
            PseCo~\cite{PseCo} & box & - & - & $19.81$ & $38.64$  \\
            MAFEA~(Ours) & box & $\textbf{12.47}$ & $\textbf{22.94}$ & $\textbf{6.36}$ & $\textbf{9.18}$ \\
		\hline\hline
	\end{tabular}
\end{table*}
\endgroup

\begingroup
\renewcommand{\arraystretch}{1} 
\begin{table*}[t]
	\caption{Evaluation on the FSC-147 dataset in few- and zero-shot settings.}
	\label{tab: FSC147 3-shot table}
	\centering
	\begin{tabular}{l|c|cc|cc}
		\hline\hline
            \multirow{2}{*}{Method} & \multirow{2}{*}{Exemplar} &
            \multicolumn{2}{c|}{Val} & \multicolumn{2}{c}{Test} \\
            \cline{3-6} 
            & & MAE~$\downarrow$ & RMSE~$\downarrow$ & MAE~$\downarrow$ & RMSE~$\downarrow$ 
            \\ 
            \hline \hline
            ZSCNet~\cite{ZSCNet} & class name & $26.93$ & $88.63$ & $22.09$ & $115.17$ \\
            SAM-Free~\cite{SAM-Free} & class name & - & - & $24.79$ & $137.15$ \\
            PseCo~\cite{PseCo} & class name & $23.90$ & $100.33$ & $16.58$ & $129.77$ \\
            \hline
            CounTR~\cite{CounTR} & $\times$ & $17.40$ & $70.33$ & $14.12$ & $108.01$ \\
            LOCA~\cite{LOCA} & $\times$ & $17.43$ & $54.96$ & $16.22$ & $103.96$ \\
            GCNet~\cite{GCNet} & $\times$ & $19.50$ & $63.13$ & $17.83$ & $102.89$ \\
            MAFEA~(Ours) & $\times$ & $14.54$ & $60.44$ & $13.23$ & $105.99$ \\
            \hline 
    	GMN~\cite{GMN}  & box & $29.66$ & $89.81$ & $26.52$ & $124.57$ \\
    	FamNet~\cite{FamNet}  & box & $24.32$ & $70.94$  & $22.56$ & $101.41$ \\
            CFOCNet~\cite{CFOCNet}  & box & $21.19$ & $61.41$ & $22.10$ & $112.71$ \\
            RCAC~\cite{RCAC} & box & $20.54$ & $60.78$ & $20.21$ & $81.86$ \\
            Counting-DETR~\cite{Counting-DETR} & box & - & - & $16.79$ & $123.56$ \\
            BMNet+~\cite{BMNet+} & box & $15.74$ & $58.53$ & $14.62$ & $91.83$ \\
            SPDCN~\cite{SPDCN} & box & $14.59$ & $49.97$ & $13.51$ & $96.80$ \\
            CounTR~\cite{CounTR} & box & $13.13$ & $49.83$ & $11.95$ & $91.23$ \\
            SAFECount~\cite{SAFECount} & box & $15.28$ & $47.20$ & $14.32$ & $85.54$ \\
            LOCA~\cite{LOCA} & box & $10.24$ & $32.56$ & $10.79$ & $56.97$ \\
            SAM-Free~\cite{SAM-Free} & box & - & - & $19.95$ & $132.16$ \\
            GCNet~\cite{GCNet} & box & $19.61$ & $66.22$ & $17.86$ & $106.98$ \\
            CSTrans~\cite{cstrans} & box & $18.10$ & $58.45$ & $16.38$ & $93.51$ \\
            PseCo~\cite{PseCo} & box & $15.31$ & $68.34$ & $13.05$ & $112.86$ \\
            MAFEA~(Ours) & box & $\textbf{8.92}$ & $\textbf{32.45}$ & $\textbf{9.84}$ & $\textbf{56.68}$ \\
		\hline\hline
	\end{tabular}
\end{table*}
\endgroup

\subsubsection{Configuration of Multi-Class Subset.}
Although the FSC-147 contains many objects in each image, the scene of each image is mainly composed of single-class objects.
Due to this inherent characteristic, the evaluation of the FSC-147 might not accurately assess the ability of the model to identify the target class within an image containing diverse object categories.
For a quantitative assessment of whether the model suffers from the target confusion problem, we construct a multi-class subset of the FSC-147~(FSC-147-Multi). We selectively remove images where objects from other classes amount to less than 20\% of the target class objects, to exclude single-object predominant images in the multi-class experiments.
The indices of the images that make up the FSC-147-Multi can be found in Table~\ref{tab: multi-class subset indices}, and experimental results are detailed in Sec.~4.3.

\subsubsection{Metrics.}
Generally, the counting methods are evaluated using Mean Absolute Error~(MAE) and Root Mean Squared Error~(RMSE). 
These metrics are defined as follows:
\begin{equation}
    \begin{aligned}
        \label{eq:define_metrics}
        &\text{MAE}=\frac{1}{n}\sum_{i=1}^{n}\left\vert y_i - \hat{y}_i\right\vert , \\
        &\text{RMSE}= \sqrt{\frac{1}{n}\sum_{i=1}^{n}\left(y_i - \hat{y}_i\right)^2},
    \end{aligned}
\end{equation}
where $n$ denotes the number of images, $y_i$ and $\hat{y_i}$ are the ground truth and predicted counts for $i$-th image, respectively.

\input{fig/figure3}

\subsection{Comparison with the State-of-the-Art}
To evaluate the model's robustness against target confusion, assessments need to be conducted in multi-class scenes where diverse class objects coexist.
Given that the FSC-147 dataset mainly consists of simple scenes with only target class objects, its limited multi-class scenes are not suitable for target confusion evaluation.
Thus, we construct a multi-class subset of the FSC-147~(FSC-147-Multi), where images contain non-target objects more than $20~\%$ of the number of existing target class objects.
Initially, we compare our model with state-of-the-art~(SOTA) methods in a complex multi-class scenario and then extend the evaluation to a simpler single-class scenario. 

\subsubsection{Evaluation on a multi-class scenario}
\label{multi}
We compare our method with SOTA methods on multi-class datasets: FSCD-LVIS and FSC-147-Multi, as shown in Table~\ref{tab: Multi-scenario table}.
On the FSCD-LVIS, our method outperforms all baselines, showing an improvement of $11~\%$ in MAE and $4.6~\%$ in RMSE compared to the second-best performer.
On the FSC-147-Multi, our results demonstrate a significant performance gap compared to the SOTA methods in both zero- and few-shot settings.
In the zero-shot setting, methods can be categorized into two types: exemplar-based~(class name) and exemplar-free~($\times$).
Exemplar-free methods, including ours, do not receive target class information and only count dominant class objects in the image.
Despite this, they outperform exemplar-based methods, indicating that existing zero-shot methods face challenges with target confusion.
In the few-shot setting, our method shows exceptional performance, thanks to its effective addressing of target confusion.

In Fig.~\ref{fig:visualization}, the 1st and 2nd rows illustrate qualitative results of the FSCD-LVIS. 
As shown in the 3rd to 6th columns, prior methods struggle to identify the target class in a query using exemplar images; they often miscount non-target objects that share similar scale or appearance with the target.
In Fig.~\ref{fig:add_visualize}, we visualize results when different exemplars are provided in a multi-class scene. Despite the distinct appearances of the exemplar images, previous methods suffer from target confusion, localizing all object-like areas in the query image, regardless of the exemplar provided.
In contrast, our method excels in precisely distinguishing target objects based on the exemplar images, a success attributable to our mutually-aware feature learning.

\subsubsection{Evaluation on a single-class scenario}
We further evaluate our method on validation and test sets of FSC-147, as shown in Table~\ref{tab: FSC147 3-shot table}. 
Our method surpasses all baselines in both zero-shot and few-shot settings, demonstrating the effectiveness of MAFEA, even in a single-class scenario. As shown in Fig.~\ref{fig:visualization}, the 3rd and 4th rows show the results for the FSC-147.
Our method makes more accurate predictions compared to other methods, especially on dense images.

\begingroup
\setlength{\tabcolsep}{5pt} 
\begin{table}[t]
\caption{Cross-dataset generalization performance on CARPK.}
\label{tab: cross-dataset}
\begin{tabular}{l|cc}
        \hline\hline
            \multirow{2}{*}{Method} & 
            \multicolumn{2}{c}{CARPK} \\
            \cline{2-3} & MAE~$\downarrow$ & RMSE~$\downarrow$ \\
            \hline \hline
            GMN~\cite{GMN} & $32.92$ & $39.88$ \\ 
            FamNet~\cite{FamNet} & $28.84$ & $44.47$ \\
            RCAC~\cite{RCAC} & $17.98$ & $24.21$ \\
            BMNet+~\cite{BMNet+}  & $10.44$ & $13.77$ \\
            SPDCN~\cite{SPDCN} & $18.15$ & $21.61$ \\
            SAFECount~\cite{SAFECount} & $16.66$ & $24.08$ \\
            LOCA~\cite{LOCA} & $9.97$ & $12.51$ \\
            CSTrans~\cite{cstrans} & $20.84$ & $24.64$ \\
            MAFEA~(Ours) & $\boldsymbol{9.19}$ & $\boldsymbol{11.90}$ \\
            \hline\hline
    \end{tabular}
\end{table}
\endgroup

\input{fig/figure4}

\subsection{Cross-Dataset Generalization}
We evaluate the generalization ability of our model on a car counting dataset, CARPK. 
To avoid overlap between the train and test sets, the tested models are pre-trained with FSC-147 by excluding its car category.
The results are summarized in Table~\ref{tab: cross-dataset}.
Note that we do not fine-tune our model on the CARPK.
As reported, our method outperforms the current state-of-the-art methods.
It demonstrates the robustness of our method in cross-dataset generalization.

\subsection{Ablation Study}
To verify the effectiveness of our approach, we conduct extensive ablation studies on FSCD-LVIS and FSC-147.
Specifically, we begin with component-level analysis on multi-class and single-class scenarios and then investigate the impact of the number of exemplars. 
Additionally, we visualize the attention maps of the Alignment Score~(AS) map to validate the role of the background token.

\begingroup
\setlength{\tabcolsep}{2.5pt} 
\begin{table}[t]
	\caption{Ablation study on Mutual Relation Modeling~(MRM), Background Token~(BT), and Target-Background Discriminitive~(TBD) Loss in FSCD-LVIS. `ALL' denotes the performance on the entire area of the image. `Target' denotes performance within the area encompassing all target objects, and `Non-Target' signifies performance in the complementary area.}
	\label{tab: fscd_lvis component-level ablation}
	\centering
	\begin{tabular}{ccc|cc|cc|cc}
		\hline\hline
            \multirow{2}{*}{MRM} &
            \multirow{2}{*}{BT} &
            \multirow{2}{*}{TBD} &
            \multicolumn{2}{c|}{ALL} &
            \multicolumn{2}{c|}{Target} &
            \multicolumn{2}{c}{Non-Target} \\
            \cline{4-9} & & & MAE & RMSE & MAE & RMSE & MAE & RMSE  \\
            \hline \hline
            $\cdot$ & $\cdot$  & $\cdot$ & $14.89$ & $25.95$ & $18.64$ & $28.84$ & $15.65$ & $23.52$ \\
            \checkmark & $\cdot$ & $\cdot$  & $13.90$ & $25.04$ & $13.88$ & $24.33$ & $10.12$ & $18.22$  \\
            \checkmark & \checkmark & $\cdot$ & $13.28$ & $24.13$ & $13.69$ & $24.26$ & $9.91$ & $16.42$  \\
            \checkmark & \checkmark & \checkmark & $12.47$ & $22.94$ & $12.70$ & $22.62$ & $7.54$ & $14.88$ \\
            \hline\hline
	\end{tabular}
\end{table}
\endgroup

\begingroup
\setlength{\tabcolsep}{2.5pt} 
\begin{table}[t]
	\caption{Ablation study on Mutual Relation Modeling~(MRM), Background Token~(BT), and Target-Background Discriminitive~(TBD) Loss in FSC-147-Multi.}
	\label{tab: multi-class subset component-level ablation}
	\centering
	\begin{tabular}{ccc|cc|cc|cc}
		\hline\hline
            \multirow{2}{*}{MRM} &
            \multirow{2}{*}{BT} &
            \multirow{2}{*}{TBD} &
            \multicolumn{2}{c|}{ALL} &
            \multicolumn{2}{c|}{Target} &
            \multicolumn{2}{c}{Non-Target} \\
            \cline{4-9} & & & MAE & RMSE & MAE & RMSE & MAE & RMSE  \\
            \hline \hline
            $\cdot$ & $\cdot$ & $\cdot$ & $12.01$ & $21.81$ & $7.84$ & $15.13$ & $5.85$ & $11.13$  \\
            \checkmark & $\cdot$ & $\cdot$  & $8.26$ & $11.90$ & $5.65$ & $9.29$ & $3.34$ & $6.44$ \\
            \checkmark & \checkmark & $\cdot$ & $7.40$ & $11.39$ & $5.42$ & $8.07$ & $2.80$ & $5.97$  \\
            \checkmark & \checkmark & \checkmark & $6.36$ & $9.18$ & $4.58$ & $6.84$ & $2.60$ & $5.20$  \\
            \hline\hline
	\end{tabular}
\end{table}
\endgroup

\input{fig/figure5}

\subsubsection{Component-level analysis on multi-class scenario}
Firstly, we verify the effect of integrating mutual relation modeling into the feature extractor. 
In Table~\ref{tab: fscd_lvis component-level ablation} and Table~\ref{tab: multi-class subset component-level ablation}, the 1st row presents the result of the model that independently computes query and exemplar features by a shared feature extractor.
Compared with it, MRM shows noteworthy enhancements in both datasets, with a $6.6\%$ MAE gain in FSCD-LVIS, and a $31.2\%$ MAE improvement in FSC-147-Multi.
This highlights the importance of mutual awareness in computing query and exemplar features.
Subsequently, we delve into the impact of BT and TBD loss.
As shown in the 3rd and 4th rows, while BT yields a slight performance improvement when used alone, its combination with TBD loss leads to a notable performance enhancement.
BT brings a performance gain of $4.5\%$ MAE in FSCD-LVIS and $10.4\%$ MAE in FSC-147-Multi, while the TBD loss achieves an additional performance gain of $6.1\%$ MAE in FSCD-LVIS and $14.1\%$ in FSC-147-Multi.
It demonstrates that minimizing undesired interactions between query and exemplar features enhances target recognition of the model. 

\begingroup
\setlength{\tabcolsep}{4pt} 
\begin{table}[t]
	\caption{Ablation study on Mutual Relation Modeling~(MRM), Background Token~(BT), and Target-Background Discriminitive Loss~(TBD) in the FSC-147 dataset.}
	\label{tab: fsc147 component-level ablation}
	\centering
	\begin{tabular}{ccc|cc|cc}
		\hline\hline
            \multirow{2}{*}{MRM} &
            \multirow{2}{*}{BT} &
            \multirow{2}{*}{TBD} &
            \multicolumn{2}{c|}{Val} &
            \multicolumn{2}{c}{Test} \\
            \cline{4-7} & & & MAE~$\downarrow$ & RMSE~$\downarrow$ & MAE~$\downarrow$ & RMSE~$\downarrow$  \\
            \hline \hline
            $\cdot$ & $\cdot$  & $\cdot$ & $12.43$ & $42.97$ & $12.91$ & $77.89$ \\
            \checkmark & $\cdot$  & $\cdot$ & $10.94$ & $34.56$ & $11.43$ & $71.43$ \\
            \checkmark & \checkmark & $\cdot$ & $10.04$ & $36.47$ & $10.78$ & $88.43$ \\
            \checkmark & \checkmark & \checkmark & $8.92$ & $32.45$ & $9.84$ & $56.68$ \\
            \hline\hline
	\end{tabular}
\end{table}
\endgroup

Furthermore, we assess performance in the target and non-target regions to verify whether the models count only the target objects. This is imperative since the evaluation within the entire region may compensate for potential under-predictions in the target region by incorrect predictions in the non-target region. 
In the few-shot object counting~(FSOC), each object is annotated only with its center point, which is insufficient to estimate good boundaries between the target and non-target region.
To overcome this limitation, we expand each point annotation to encompass an area equivalent to the maximum size of exemplars. Consequently, the target region encompasses all target objects, while the non-target region covers the complementary area.
We provide the visualization of the target region map and corresponding results in Figure~\ref{fig:supple_target_fig}.
If there is no target confusion issue, the model’s prediction in the non-target region should be zero. 
Impressively, MRM, BT, and TBD bring substantial performance improvements in both target and non-target areas.
The notable enhancement in the non-target area validates that the proposed components alleviate target confusion as intended.

\begingroup
\setlength{\tabcolsep}{1.5pt} 
\begin{table}[t]
	\caption{Impact of the number of exemplars}
	\label{tab: impact of exemplar size}
	\centering
	\begin{tabular}{c|cc|cc|cc|cc}
		\hline\hline
            \multirow{3}{*}{Shot} &
            \multicolumn{2}{c|}{FSCD-LVIS} &
            \multicolumn{6}{c}{FSC-147} \\
            \cline{2-9} &
            \multicolumn{2}{c|}{Test} & \multicolumn{2}{c|}{Multi} & \multicolumn{2}{c|}{Val} & \multicolumn{2}{c}{Test} \\
            \cline{2-9} & MAE & RMSE & MAE & RMSE & MAE & RMSE & MAE & RMSE \\
            \hline \hline
            $0$ & $14.72$ & $27.42$ & $12.12$ & $26.51$ & $14.54$ & $60.44$ & $13.23$ & $105.99$ \\
            $1$ & $13.30$ & $24.26$ & $8.32$ & $18.80$ & $11.14$ & $42.37$ & $11.68$ & $99.17$ \\
            $2$ & $12.81$ & $23.68$ & $6.52$ & $15.32$ & $10.79$ & $42.04$ & $10.91$ & $80.74$ \\
            $3$ & $12.47$ & $22.94$ & $6.36$ & $9.18$ & $8.92$ & $32.45$ & $9.84$ & $56.68$ \\
            \hline\hline
	\end{tabular}
\end{table}
\endgroup

\subsubsection{Component-level analysis on single-class scenario}
We further assess our proposed module in the validation and test sets of FSC-147.
As shown in Table~\ref{tab: fsc147 component-level ablation}, MRM achieves a performance gain of $12.0\%$ MAE and $11.5\%$ MAE for the validation and test sets, respectively. The ablation of BT brings an improvement of $8.2\%$ MAE and $5.7\%$ MAE for the validation and test sets, respectively. Also, TBD brings an improvement in MAEs of $11.2\%$ and $8.7\%$ for validation and test sets, respectively. These results confirm the effectiveness of our method even in the single-class scenario.

\subsubsection{The number of exemplars}
We further investigate the impact of the exemplar's numbers in Table~\ref{tab: impact of exemplar size}.
From the zero-shot where exemplar features are replaced by learnable tokens, to the 3-shot, our method provides better performance as more exemplars are provided.
These results validate that the exemplar features contribute to refining the query features in a mutually-aware manner.

\input{fig/figure6}

\subsubsection{Influence of the background token}
To confirm that BT highlights the background region, including non-target objects, we visualize the alignment score~(AS) map in Fig.~\ref{fig:ASmap_figure}.
AS map is obtained by summing up the alignment scores of BT or exemplar features.
BT alone fails to activate the non-target objects because there is no mechanism enforcing its interaction with the query's background.
Our TBD loss directly aligns exemplars and BT with the query's target and non-target features.
As depicted in Fig.~\ref{fig:ASmap_figure}, BT operates as intended only when used with TBD loss.

Additionally, we conduct an ablation study on the number of BTs in Fig.~\ref{fig:background token}.
While removing BT leads to degraded performance, the quantity of tokens has a minor impact on the results.
BT aims to prevent undesired interactions between the query's non-target features and the exemplar features.
Therefore, the presence of the BT, rather than its quantity, truly matters.

\subsection{Complexity Comparison}
We compare the computational complexity and the number of parameters in Table~\ref{tab: complexity}. Although our model uses more parameters than CNN-based models, its computational cost is similar or lower. Compared to ViT-based models like CounTR, our model uses fewer parameters.

\subsection{Additional Qualitative Results}
We present more qualitative results. As shown in Figure~\ref{fig:supple_qualitative_success}, our model accurately predicts counts across sparse and dense scenes for novel classes. However, our method sometimes confuses the target objects when non-target objects have a similar appearance or size to the target class, as seen in the 1st row of Figure~\ref{fig:supple_qualitative_failure}. Moreover, our method may mistakenly count both classes when the exemplar image contains multiple classes, as shown in the 2nd row. Additionally, in the 3rd to 5th rows, our model struggles with images that contain incredibly high object densities, especially those with more than 1000 instances.

\input{fig/figure7}

\begingroup
\begin{table}[t]
\setlength{\tabcolsep}{4pt} 
    \caption{Computational complexity and the number of parameters.}
    \label{tab: complexity}
    \centering
    \begin{tabular}{l|c|c|c}
        \hline\hline
            Method & Encoder & Param~(M) & GFLOPS~(G) \\
            \hline \hline
            SAFECount & CNN & $32$ & $366$ \\
            LOCA & CNN & $35$ & $80$ \\
            CounTR & ViT & $100$ & $91$ \\
            MAFEA~(Ours) & ViT & $93$ & $122$ \\
            \hline\hline
    \end{tabular}
\end{table}
\endgroup

\input{fig/figure8}

%% file: fig/figure3.tex
\begin{figure*}[t!]
    \centering
    \includegraphics[width=1.\textwidth]{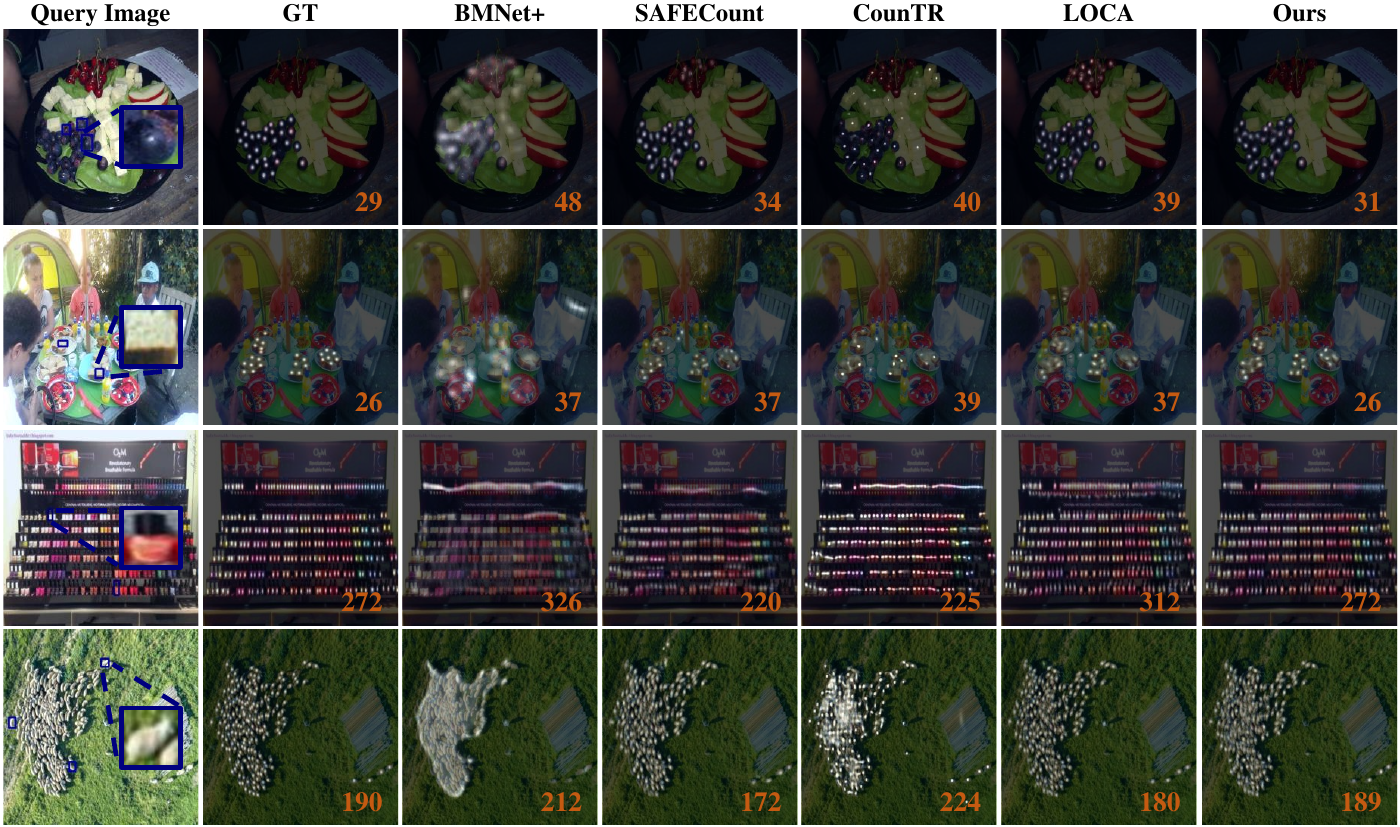}
    \caption{
    Qualitative results: 1st and 2nd rows from FSCD-LVIS dataset, and the 3rd and 4th rows from FSC-147 dataset.
    Each box in the query image is a box annotation for an exemplar, while the numbers in the pictures are the counting results. Best viewed with zoom-in.
    }
    \label{fig:visualization}
\end{figure*}

%% file: fig/figure4.tex
\begin{figure*}[t]
    \centering
    \includegraphics[width=1.\textwidth]{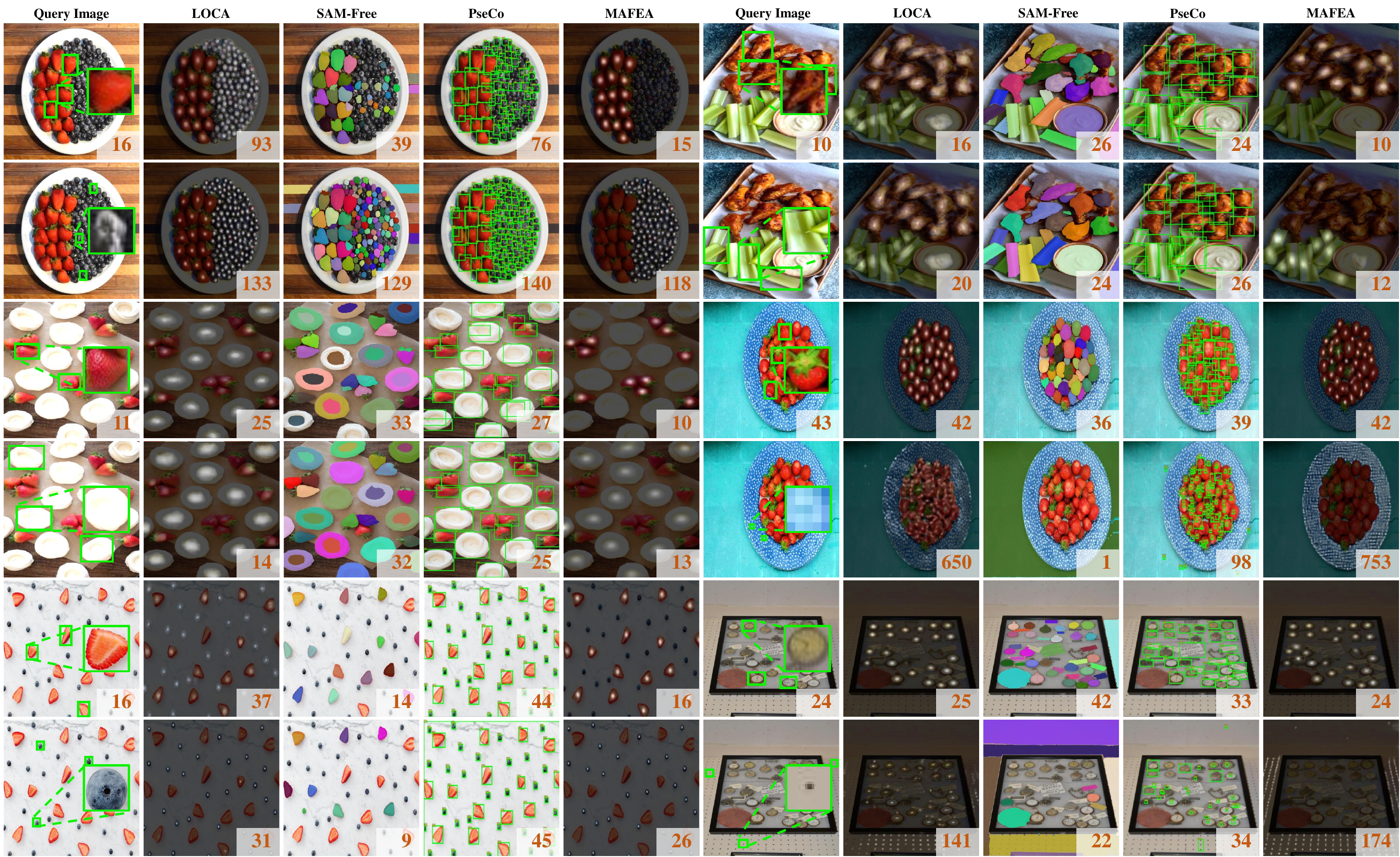}
    \caption{
    Qualitative results for a multi-class scenario.
    Each box in the query image is a box annotation for an exemplar, while the numbers in the pictures are the counting results. Best viewed with zoom-in.
    }
    \label{fig:add_visualize}
\end{figure*}

%% file: fig/figure5.tex
\begin{figure*}[t]
    \centering
    \includegraphics[width=1.\textwidth]{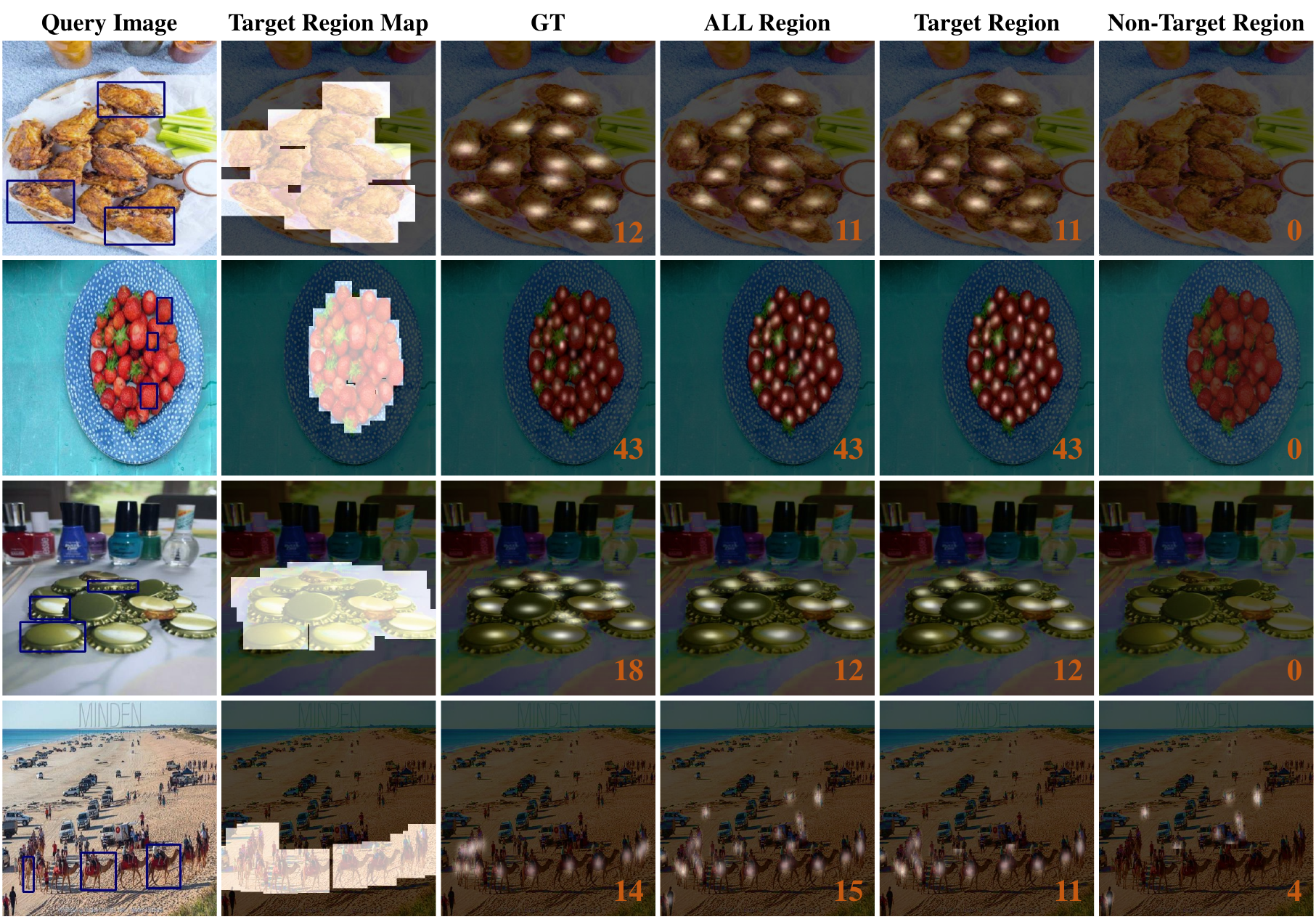}
    \caption{
        Results in multi-class scenes within the FSC-147 dataset. From left to right: query image, target region map, ground-truth density map, our prediction on all region, target region, and non-target region. Each box in the query image is a box annotation for each exemplar image, while the numbers in images are the counting results.
    }
    \label{fig:supple_target_fig}
\end{figure*}

%% file: fig/figure6.tex
\begin{figure}[t]
    \centering
    \includegraphics[width=1.\columnwidth]{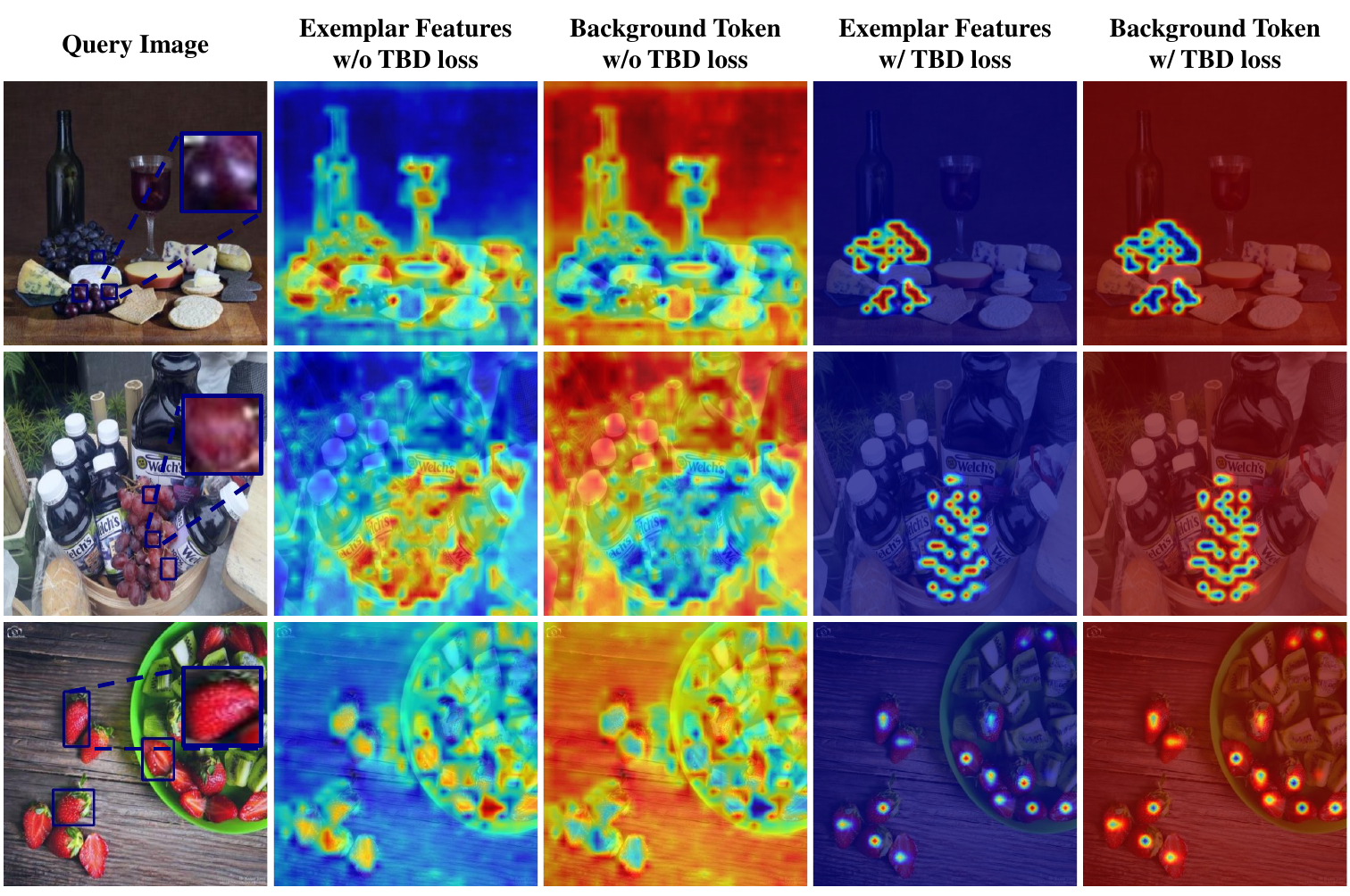}
    \caption{
        Alignment Score~(AS) map of exemplar features and background token. 2nd and 3rd columns are from model trained w/o TBD loss, and the 4th and 5th columns are from model trained w/TBD loss.
    }
    \label{fig:ASmap_figure}
\end{figure}

%% file: fig/figure7.tex
\begin{figure}[t]
    \centering
    \includegraphics[width=0.9\columnwidth]{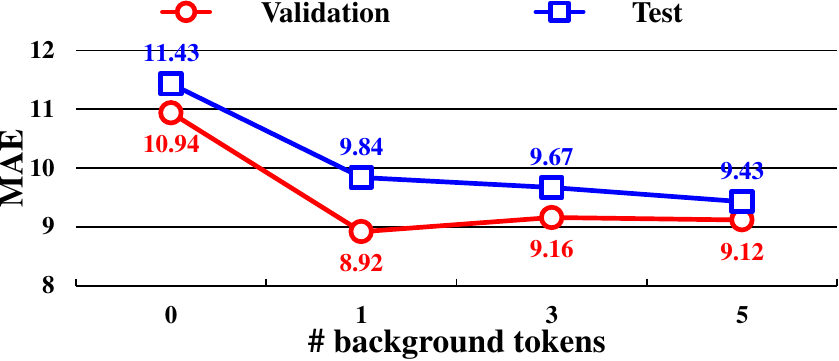}
    \caption{
        Results w.r.t the number of background tokens.
    }
    \label{fig:background token}
\end{figure}

%% file: fig/figure8.tex
\begin{figure*}[t]
\begingroup
    \begin{subfigure}[t]{0.49\textwidth}
        \centering
            \includegraphics[width=0.95\columnwidth]{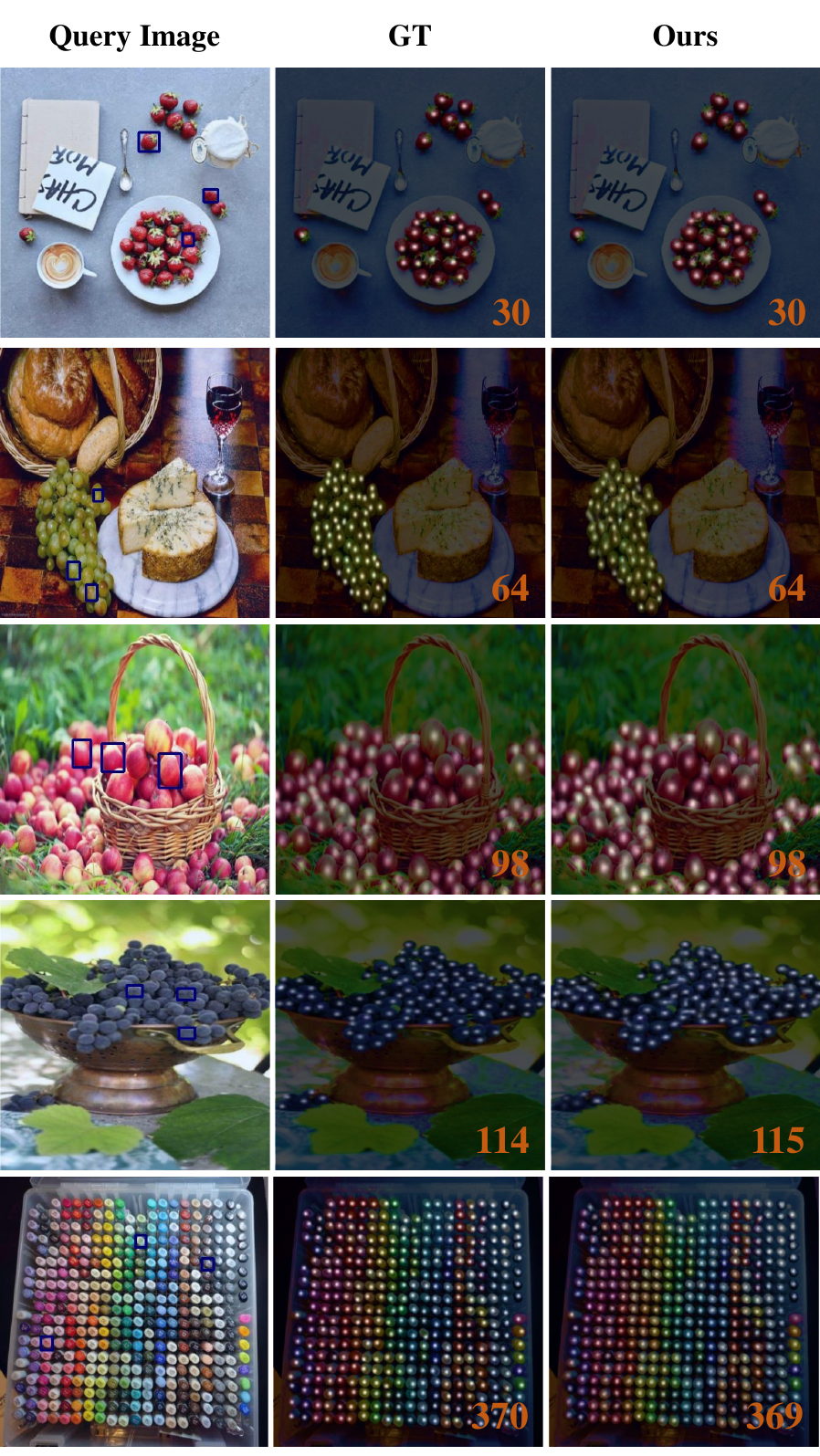}
            \caption{
                Success cases
            }
            \label{fig:supple_qualitative_success}
    \end{subfigure}
\endgroup
\hfill
\begingroup
    \begin{subfigure}[t]{0.49\textwidth}
        \centering
            \includegraphics[width=0.95\columnwidth]{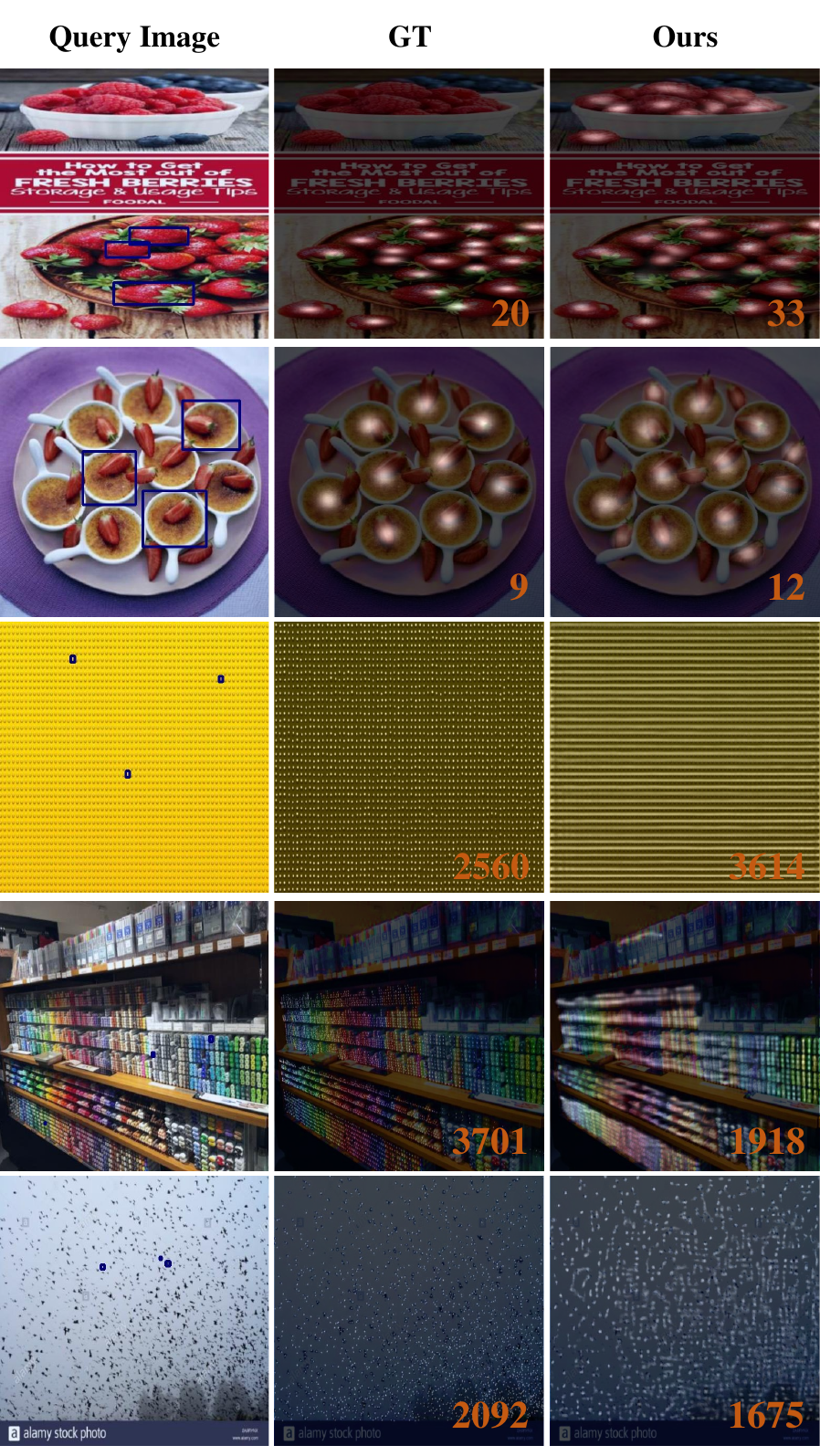}
            \caption{
                Failure cases
            }
            \label{fig:supple_qualitative_failure}
    \end{subfigure}
\endgroup
\caption{Success cases and failure cases on FSC-147 dataset.}
\label{fig:supple_qualitative}
\end{figure*}

%% file: contents/5_conclusion.tex
\section{Conclusion}
\label{sec:conclusion}
In this work, we highlight and address the target confusion problem in multi-class scenarios.
We conjecture that target confusion arises because existing methods encode query and exemplar features independently, without any explicit interaction during feature extraction.
This lack of interaction causes insufficient target awareness, making it difficult for the model to distinguish between target and non-target objects in multi-class scenes.
To settle this problem, we designed the model to distinguish between target and non-target objects by producing target-aware features through early interaction in the feature extraction process. The comprehensive results validate that our proposed method effectively mitigates target confusion. 
However, as discussed in Sec.~4.7, despite its robustness, our method often fails to identify the correct target class when objects from visually similar classes co-exist.
To overcome this limitation, future work could explore fine-grained classification techniques, which are specifically designed to capture subtle differences between visually similar classes. Incorporating such techniques could improve the model’s ability to distinguish visually similar classes, thereby enhancing its real-world performance.

%% file: contents/6_appendix.tex
\section{Appendix}\label{sec:appendix}

\begin{figure*}[t]
    \centering
    \includegraphics[width=1.\textwidth]{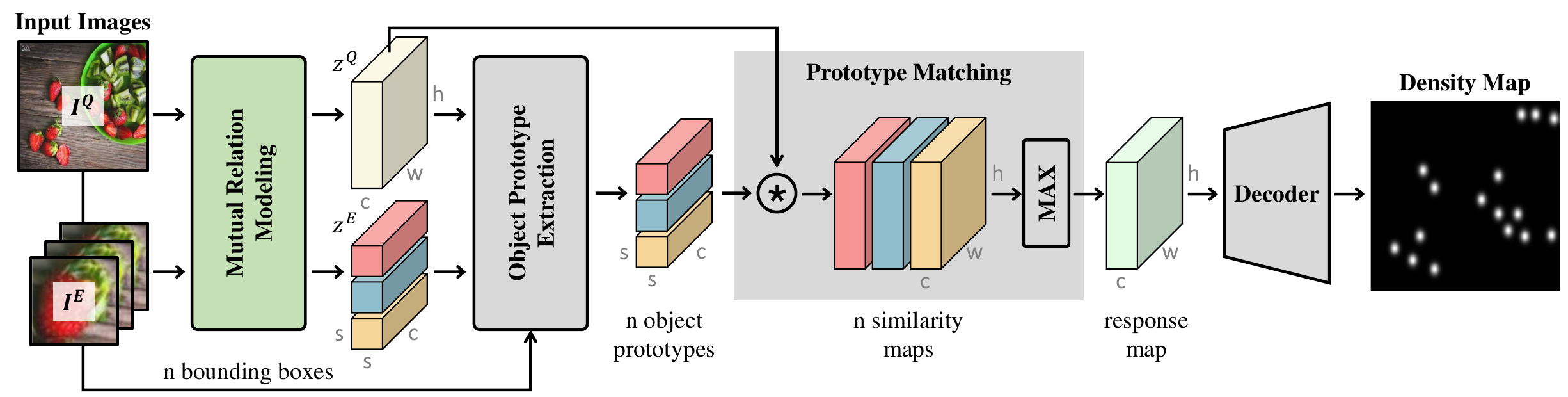}
    \caption{
    Details about the Relation Learner. Input images $I^Q$ and $I^E$ are encoded into features $z^Q$ and $z^E$. The Object Prototype Extraction~(OPE) module combines shape properties from n bounding boxes with appearance properties from exemplar features $z^E$ to generate n object prototypes. $z^Q$ is depth-wise correlated~(*) with n object prototypes, resulting in n similarity maps. The response map is then obtained by computing the per-element maximum of n similarity maps and upsampled by the decoder to the final density map. 
    }
    \label{fig:supple_fig}
\end{figure*}

\subsection{Additional Details}
As described in Sec.~3, our research focus lies in the feature extractor with Mutual Relation Modeling~(MRM).
We adopt the same relation learner and decoder as LOCA, so we do not claim contribution in this regard.
In this section, we provide details about the relation learner, as illustrated in Figure~\ref{fig:supple_fig}.
The relation learner, inspired by LOCA, consists of the Object Prototype Extraction~(OPE) and Prototype Matching~(PM) modules. 
The sole difference lies in the OPE modules, where LOCA obtains appearance features from ROI pooling in the query feature~($z^Q)$ using box annotations, while we leverage exemplar features~($z^E$) from the MRM encoder. The spatial size~(s) of exemplar features is set to $3$, consistent with LOCA.

%% file: cas-refs.bib
@CONFERENCE{CLIP,
  title={Learning transferable visual models from natural language supervision},
  author={Radford, Alec and Kim, Jong Wook and Hallacy, Chris and Ramesh, Aditya and Goh, Gabriel and Agarwal, Sandhini and Sastry, Girish and Askell, Amanda and Mishkin, Pamela and Clark, Jack and others},
  booktitle={International conference on machine learning},
  pages={8748--8763},
  year={2021},
  organization={PMLR}
}

@CONFERENCE{fewshotseg,
  title={Learning self-target knowledge for few-shot segmentation},
  author={Chen, Yadang and Chen, Sihan and Yang, Zhi-Xin and Wu, Enhua},
  journal={Pattern Recognition},
  volume={149},
  pages={110266},
  year={2024},
  publisher={Elsevier}
}

@CONFERENCE{SAM,
  title={Segment anything},
  author={Kirillov, Alexander and Mintun, Eric and Ravi, Nikhila and Mao, Hanzi and Rolland, Chloe and Gustafson, Laura and Xiao, Tete and Whitehead, Spencer and Berg, Alexander C and Lo, Wan-Yen and others},
  booktitle={Proceedings of the IEEE/CVF International Conference on Computer Vision},
  pages={4015--4026},
  year={2023}
}

@CONFERENCE{PseCo,
  title={Point Segment and Count: A Generalized Framework for Object Counting},
  author={Huang, Zhizhong and Dai, Mingliang and Zhang, Yi and Zhang, Junping and Shan, Hongming},
  booktitle={Proceedings of the IEEE/CVF Conference on Computer Vision and Pattern Recognition},
  pages={17067--17076},
  year={2024}
}

@CONFERENCE{SAM-Free,
  title={Training-free object counting with prompts},
  author={Shi, Zenglin and Sun, Ying and Zhang, Mengmi},
  booktitle={Proceedings of the IEEE/CVF Winter Conference on Applications of Computer Vision},
  pages={323--331},
  year={2024}
}

@CONFERENCE{ZSCNet,
  title={Zero-shot object counting},
  author={Xu, Jingyi and Le, Hieu and Nguyen, Vu and Ranjan, Viresh and Samaras, Dimitris},
  booktitle={Proceedings of the IEEE/CVF Conference on Computer Vision and Pattern Recognition},
  pages={15548--15557},
  year={2023}
}

@CONFERENCE{GCNet,
  title={GCNet: Probing self-similarity learning for generalized counting network},
  author={Wang, Mingjie and Li, Yande and Zhou, Jun and Taylor, Graham W and Gong, Minglun},
  journal={Pattern Recognition},
  volume={153},
  pages={110513},
  year={2024},
  publisher={Elsevier}
}

@CONFERENCE{cstrans,
  title={CSTrans: Correlation-guided Self-Activation transformer for counting everything},
  author={Gao, Bin-Bin and Huang, Zhongyi},
  journal={Pattern Recognition},
  pages={110556},
  year={2024},
  publisher={Elsevier}
}

@CONFERENCE{car2,
  title={Dense center-direction regression for object counting and localization with point supervision},
  author={Tabernik, Domen and Muhovi{\v{c}}, Jon and Sko{\v{c}}aj, Danijel},
  journal={Pattern Recognition},
  volume={153},
  pages={110540},
  year={2024},
  publisher={Elsevier}
}

@CONFERENCE{macrowd,
  title={Crowd counting from single images using recursive multi-pathway zooming and foreground enhancement},
  author={Ma, Junjie and Dai, Yaping and Jia, Zhiyang and Sun, Fuchun and Tan, Yap-Peng and Liu, Jun},
  journal={Pattern Recognition},
  volume={141},
  pages={109585},
  year={2023},
  publisher={Elsevier}
}

@CONFERENCE{STNet,
  title={STNet: Scale tree network with multi-level auxiliator for crowd counting},
  author={Wang, Mingjie and Cai, Hao and Han, Xian-Feng and Zhou, Jun and Gong, Minglun},
  journal={IEEE Transactions on Multimedia},
  volume={25},
  pages={2074--2084},
  year={2022},
  publisher={IEEE}
}

@CONFERENCE{Interlayer,
  title={Interlayer and intralayer scale aggregation for scale-invariant crowd counting},
  author={Wang, Mingjie and Cai, Hao and Zhou, Jun and Gong, Minglun},
  journal={Neurocomputing},
  volume={441},
  pages={128--137},
  year={2021},
  publisher={Elsevier}
}

@CONFERENCE{MAE,
  title={Masked autoencoders are scalable vision learners},
  author={He, Kaiming and Chen, Xinlei and Xie, Saining and Li, Yanghao and Doll{\'a}r, Piotr and Girshick, Ross},
  booktitle={Proceedings of the IEEE/CVF conference on computer vision and pattern recognition},
  pages={16000--16009},
  year={2022}
}

@CONFERENCE{LOCA,
  title={A Low-Shot Object Counting Network With Iterative Prototype Adaptation},
  author={Djuki{\'c}, Nikola and Luke{\v{z}}i{\v{c}}, Alan and Zavrtanik, Vitjan and Kristan, Matej},
  booktitle={Proceedings of the IEEE/CVF International Conference on Computer Vision},
  pages={18872--18881},
  year={2023}
}

@CONFERENCE{class-specific,
  title={Pedestrian detection in crowded scenes},
  author={Leibe, Bastian and Seemann, Edgar and Schiele, Bernt},
  booktitle={2005 IEEE computer society conference on computer vision and pattern recognition (CVPR'05)},
  volume={1},
  pages={878--885},
  year={2005},
  organization={IEEE}
}

@CONFERENCE{class-specific2,
  title={Automatic adaptation of a generic pedestrian detector to a specific traffic scene},
  author={Wang, Meng and Wang, Xiaogang},
  booktitle={CVPR 2011},
  pages={3401--3408},
  year={2011},
  organization={IEEE}
}

@CONFERENCE{class-specific3,
  title={End-to-end people detection in crowded scenes},
  author={Stewart, Russell and Andriluka, Mykhaylo and Ng, Andrew Y},
  booktitle={Proceedings of the IEEE conference on computer vision and pattern recognition},
  pages={2325--2333},
  year={2016}
}

@CONFERENCE{class-specific-regression,
  title={Perspective-guided convolution networks for crowd counting},
  author={Yan, Zhaoyi and Yuan, Yuchen and Zuo, Wangmeng and Tan, Xiao and Wang, Yezhen and Wen, Shilei and Ding, Errui},
  booktitle={Proceedings of the IEEE/CVF international conference on computer vision},
  pages={952--961},
  year={2019}
}

@CONFERENCE{class-specific-regression2,
  title={Single-image crowd counting via multi-column convolutional neural network},
  author={Zhang, Yingying and Zhou, Desen and Chen, Siqin and Gao, Shenghua and Ma, Yi},
  booktitle={Proceedings of the IEEE conference on computer vision and pattern recognition},
  pages={589--597},
  year={2016}
}

@CONFERENCE{CARPK,
  title={Drone-based object counting by spatially regularized regional proposal network},
  author={Hsieh, Meng-Ru and Lin, Yen-Liang and Hsu, Winston H},
  booktitle={Proceedings of the IEEE international conference on computer vision},
  pages={4145--4153},
  year={2017}
}

@CONFERENCE{animal,
  title={Counting in the wild},
  author={Arteta, Carlos and Lempitsky, Victor and Zisserman, Andrew},
  booktitle={Computer Vision--ECCV 2016: 14th European Conference, Amsterdam, The Netherlands, October 11--14, 2016, Proceedings, Part VII 14},
  pages={483--498},
  year={2016},
  organization={Springer}
}

@CONFERENCE{GMN,
  title={Class-agnostic counting},
  author={Lu, Erika and Xie, Weidi and Zisserman, Andrew},
  booktitle={Computer Vision--ACCV 2018: 14th Asian Conference on Computer Vision, Perth, Australia, December 2--6, 2018, Revised Selected Papers, Part III 14},
  pages={669--684},
  year={2019},
  organization={Springer}
}

@CONFERENCE{CFOCNet,
  title={Class-agnostic few-shot object counting},
  author={Yang, Shuo-Diao and Su, Hung-Ting and Hsu, Winston H and Chen, Wen-Chin},
  booktitle={Proceedings of the IEEE/CVF Winter Conference on Applications of Computer Vision},
  pages={870--878},
  year={2021}
}

@CONFERENCE{FamNet,
  title={Learning to count everything},
  author={Ranjan, Viresh and Sharma, Udbhav and Nguyen, Thu and Hoai, Minh},
  booktitle={Proceedings of the IEEE/CVF Conference on Computer Vision and Pattern Recognition},
  pages={3394--3403},
  year={2021}
}

@CONFERENCE{BMNet+,
  title={Represent, compare, and learn: A similarity-aware framework for class-agnostic counting},
  author={Shi, Min and Lu, Hao and Feng, Chen and Liu, Chengxin and Cao, Zhiguo},
  booktitle={Proceedings of the IEEE/CVF Conference on Computer Vision and Pattern Recognition},
  pages={9529--9538},
  year={2022}
}

@CONFERENCE{SPDCN,
  title={Scale-Prior Deformable Convolution for Exemplar-Guided Class-Agnostic Counting.},
  author={Lin, Wei and Yang, Kunlin and Ma, Xinzhu and Gao, Junyu and Liu, Lingbo and Liu, Shinan and Hou, Jun and Yi, Shuai and Chan, Antoni B},
  booktitle={British Machine Vision Conference},
  pages={313},
  year={2022}
}

@CONFERENCE{CounTR,
  title={Countr: Transformer-based generalised visual counting},
  author={Liu, Chang and Zhong, Yujie and Zisserman, Andrew and Xie, Weidi},
  booktitle={British Machine Vision Conference},
  pages     = {370},
  year      = {2022}
}

@CONFERENCE{SAFECount,
  title={Few-shot object counting with similarity-aware feature enhancement},
  author={You, Zhiyuan and Yang, Kai and Luo, Wenhan and Lu, Xin and Cui, Lei and Le, Xinyi},
  booktitle={Proceedings of the IEEE/CVF Winter Conference on Applications of Computer Vision},
  pages={6315--6324},
  year={2023}
}

@CONFERENCE{Counting-DETR,
  title={Few-shot object counting and detection},
  author={Nguyen, Thanh and Pham, Chau and Nguyen, Khoi and Hoai, Minh},
  booktitle={European Conference on Computer Vision},
  pages={348--365},
  year={2022},
  organization={Springer}
}

@CONFERENCE{RCAC,
  title={Class-agnostic object counting robust to intraclass diversity},
  author={Gong, Shenjian and Zhang, Shanshan and Yang, Jian and Dai, Dengxin and Schiele, Bernt},
  booktitle={European Conference on Computer Vision},
  pages={388--403},
  year={2022},
  organization={Springer}
}
